%% file: distance-based-bias-ppsn2012.tex
\documentclass[11pt]{article}

\usepackage{amsmath}
\usepackage{url}
\usepackage{paralist}
\usepackage{multirow}
\usepackage{color}
\usepackage{epsfig}

\usepackage{booktabs}
\usepackage{subfig}
\usepackage{url}
\usepackage{floatrow,fr-subfig}

\definecolor{myblue}{rgb}{0.165,0.34,0.5}

\title{Transfer Learning, Soft Distance-Based Bias, and\\the Hierarchical BOA}

\newcommand{\titlestringtwo}{Transfer Learning, Soft Distance-Based Bias, and the Hierarchical BOA}
\newcommand{\shortauthors}{Martin Pelikan, Mark W. Hauschild, and Pier Luca Lanzi}
\newcommand{\reportnumber}{2012004}
\newcommand{\datestring}{March 2012}

\newcommand{\mysection}[1]{\section{#1}}
\newcommand{\mysubsection}[1]{\subsection{#1}}

\date{March 27, 2012}

\author{Martin Pelikan 
\and
Mark W. Hauschild 
\and
Pier Luca Lanzi}

\begin{document}

\begin{titlepage}
\setlength{\parindent}{0pt}

\noindent
\includegraphics[width=5in]{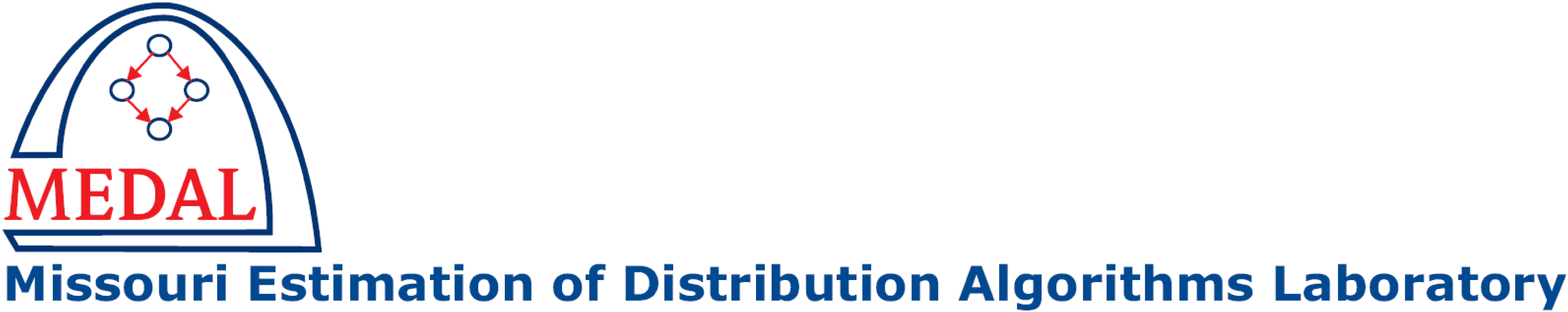}
\vspace*{0.075in}
{\color{myblue}
\hrule height 2pt
}
\vspace*{0.5in}

{\bf
\textsf{{\large
\titlestringtwo}}
}

\vspace*{0.25in}

\textsf{\shortauthors}

\vspace*{0.25in}

\textsf{MEDAL Report No. \reportnumber}

\vspace*{0.25in}

\textsf{\datestring}

\vspace*{0.25in}

{\bf \textsf{Abstract}}  

\vspace*{0.075in}

{\small \textsf{An automated technique has recently been proposed to transfer learning in the hierarchical Bayesian optimization algorithm (hBOA) based on distance-based statistics. The technique enables practitioners to improve hBOA efficiency by collecting statistics from probabilistic models obtained in previous hBOA runs and using the obtained statistics to bias future hBOA runs on similar problems. The purpose of this paper is threefold: (1)~test the technique on several classes of NP-complete problems, including MAXSAT, spin glasses and minimum vertex cover; (2)~demonstrate that the technique is effective even when previous runs were done on problems of different size; (3)~provide empirical evidence that combining transfer learning with other efficiency enhancement techniques can often yield nearly multiplicative speedups.}}

\vspace*{0.25in}

{\bf \textsf{Keywords}}

\vspace*{0.075in}
{\small \textsf{Transfer learning, inductive transfer, learning from experience, estimation of distribution algorithms, hierarchical Bayesian optimization algorithm, decomposable problems, efficiency enhancement.}}

\vfill

\noindent
\begin{minipage}{6in}
{\small \textsf{Missouri Estimation of Distribution Algorithms Laboratory (MEDAL)\\
Department of Mathematics and Computer Science, 321 ESH\\
University of Missouri--St. Louis\\
One University Blvd.,
St. Louis, MO 63121\\
E-mail: \url{medal@medal-lab.org}\\
WWW: \url{http://medal-lab.org/}\\}}
\end{minipage}

\end{titlepage}

\author{
{\bf Martin Pelikan}\\
Missouri Estimation of Distribution Algorithms Laboratory (MEDAL)\\
Dept. of Mathematics and Computer Science, 320 ESH\\
University of Missouri in St. Louis\\
One University Blvd., St. Louis, MO 63121\\
\url{martin@martinpelikan.net}\\
\url{http://martinpelikan.net/}
\and
{\bf Mark W. Hauschild}\\
Missouri Estimation of Distribution Algorithms Laboratory (MEDAL)\\
Dept. of Mathematics and Computer Science, 321 ESH\\
University of Missouri in St. Louis\\
One University Blvd., St. Louis, MO 63121\\
\url{mwh308@umsl.edu}
\and
{\bf Pier Luca Lanzi}\\
Dipartimento di Elettronica e Informazione\\
Politecnico di Milano\\
Piazza Leonardo da Vinci, 32\\
I-20133 Milano, Italy\\
\url{pierluca.lanzi@polimi.it}
}

\maketitle

\begin{abstract}
An automated technique has recently been proposed to transfer learning in the hierarchical Bayesian optimization algorithm (hBOA) based on distance-based statistics. The technique enables practitioners to improve hBOA efficiency by collecting statistics from probabilistic models obtained in previous hBOA runs and using the obtained statistics to bias future hBOA runs on similar problems. The purpose of this paper is threefold: (1)~test the technique on several classes of NP-complete problems, including MAXSAT, spin glasses and minimum vertex cover; (2)~demonstrate that the technique is effective even when previous runs were done on problems of different size; (3)~provide empirical evidence that combining transfer learning with other efficiency enhancement techniques can often yield nearly multiplicative speedups.
\end{abstract}


\mysection{Introduction}
Estimation of distribution algorithms (EDAs)~\cite{Hauschild:11c,Larranaga:02,Pelikan:02,Pelikan:12b} guide the search for the optimum by building and sampling probabilistic models of candidate solutions. The use of probabilistic models in EDAs provides a basis for incorporating prior knowledge about the problem and learning from previous runs in order to solve new problem instances of similar type with increased speed, accuracy and reliability~\cite{Pelikan:book,Hauschild:12}. However, much prior work in this area was based on hand-crafted constraints on probabilistic models~\cite{Muhlenbein:99a,Muhlenbein:02,Baluja:06,Schwarz:00*} which may be difficult to design or even detrimental to EDA efficiency and scalability~\cite{Hauschild:09c}. Recently, Pelikan and Hauschild~\cite{Pelikan:12} proposed an automated technique capable of learning from previous runs of  the hierarchical Bayesian optimization algorithm (hBOA) in order to improve efficiency of future hBOA runs on problems of similar type. The basic idea of the approach was to (1)~design a distance metric on problem variables that correlates with the expected strength of dependencies between the variables, (2)~collect statistics on hBOA models with respect to the values of the distance metric, and (3)~use the collected statistics to bias model building in hBOA when solving future problem instances of similar type. While the distance metric is strongly related to the problem being solved, the aforementioned study~\cite{Pelikan:12} described a rather general metric that can be applied to practically any problem with the objective function represented by an additively decomposable function. However, the prior study~\cite{Pelikan:12}  evaluated the proposed technique on only two classes of problems and it did not demonstrate several key features of this technique.

\begin{sloppy}
The purpose of this paper is threefold: (1)~Demonstrate the technique from ref.~\cite{Pelikan:12} on other classes of challenging optimization problems, (2)~demonstrate the ability of this technique to learn from problem instances of one size in order to introduce bias for instances of another size, and (3)~demonstrate the potential benefits of combining this technique with other efficiency enhancement techniques, such as sporadic model building~\cite{DBLP:journals/gpem/PelikanSG08}. As test problems the paper considers several classes of NP-complete additively decomposable problems, including MAXSAT, three-dimensional Ising spin glass, and minimum vertex cover. The new results together with the results published in prior work~\cite{Pelikan:12} provide strong evidence of the broad applicability and great potential of this technique for learning from experience (transfer learning) in EDAs.
\end{sloppy}

The paper is organized as follows. Section~\ref{section-hboa} outlines hBOA. Section~\ref{section-transfer-learning} discusses efficiency enhancement of estimation of distribution algorithms using inductive transfer with main focus on hBOA and the distance-based bias~\cite{Pelikan:12}. Section~\ref{section-experiments} presents and discusses experimental results. Section~\ref{section-conclusions} summarizes and concludes the paper.

\mysection{Hierarchical BOA}
\label{section-hboa}
The hierarchical Bayesian optimization algorithm (hBOA)~\cite{Pelikan:book,Pelikan:01*} works with a population of candidate solutions represented by fixed-length strings over a finite alphabet. In this paper, candidate solutions are represented by $n$-bit binary strings. The initial population of binary strings is generated at random according to the uniform distribution over candidate solutions. Each iteration starts by selecting promising solutions from the current population; here binary tournament selection without replacement is used. Next, hBOA (1)~learns a Bayesian network with local structures~\cite{Chickering:97} for the selected solutions and (2)~generates new candidate solutions by sampling the distribution encoded by the built network. To maintain useful diversity in the population, the new candidate solutions are incorporated into the original population using restricted tournament selection (RTS)~\cite{Harik:95a}. The run is terminated when termination criteria are met. In this paper, each run is terminated either when the global optimum is found or when a maximum number of iterations is reached. 

hBOA represents probabilistic models of candidate solutions by Bayesian networks with local structures~\cite{Chickering:97,Friedman:99}. A Bayesian network is defined by two components: (1)~an acyclic directed graph over problem variables specifying direct dependencies between variables and (2)~conditional probabilities specifying the probability distribution of each variable given the values of the variable's parents. A Bayesian network encodes a joint probability distribution as $p(X_1,\ldots,X_n)=\prod_{i=1}^n p(X_i|\Pi_i)$ where $X_i$ is the $i$th variable (string position) and $\Pi_i$ are the parents of $X_i$ in the underlying graph. 

To represent conditional probabilities of each variable given the variable's parents, hBOA uses decision trees~\cite{Pelikan:01*,Chickering:97}. Each internal node of a decision tree specifies a variable, and the subtrees of the node correspond to the different values of the variable. Each leaf of the decision tree for a particular variable defines the probability distribution of the variable given a condition specified by the constraints given by the path from the root of the tree to this leaf (constraints are given by the assignments of the variables along this path).

To build probabilistic models, hBOA typically uses a greedy algorithm that initializes the decision tree for each problem variable $X_i$ to a single-node tree that encodes the unconditional probability distribution of $X_i$. In each iteration, the model building algorithm tests how much a model would improve after splitting each leaf of each decision tree on each variable that is not already located on the path to the leaf. The algorithm executes the split that provides the most improvement, and the process is repeated until no more improvement is possible.
Models are evaluated using the Bayesian-Dirichlet (BDe) metric with penalty for model complexity, which estimates the goodness of a Bayesian network structure given data $D$ and background knowledge $\xi$ as $p(B|D,\xi) = c p(B|\xi) p(D|B,\xi),$
where $c$ is a normalization constant ~\cite{Chickering:97,Cooper:92}. The Bayesian-Dirichlet metric estimates the term $p(D|B,\xi)$ by combining the observed and prior statistics for relevant combinations of variables~\cite{Chickering:97}.  
To favor simpler networks to the more complex ones, the prior probability $p(B|\xi)$ is often set to decrease exponentially fast with respect to the description length of the network's parameters~\cite{Pelikan:book,Friedman:99}.

\mysection{Learning from Experience using Distance-Based Bias}
\label{section-transfer-learning}


In hBOA and other EDAs based on complex probabilistic models, building an accurate probabilistic model is crucial to the success~\cite{Larranaga:02,Pelikan:02,Hauschild:09c,Lima:11}. However, building complex probabilistic models can be  time consuming and it may require rather large populations of solutions~\cite{Larranaga:02,Pelikan:02}. That is why much effort has been put into enhancing efficiency of model building in EDAs and improving quality of EDA models even with smaller populations~\cite{Hauschild:12,Muhlenbein:02,Baluja:06,Hauschild:08,Hauschild:09b}. 
Learning from experience~\cite{Pelikan:book,Hauschild:12,Pelikan:12,Hauschild:08,Hauschild:09b} represents one approach to addressing this issue.

The basic idea of learning from experience is to gather information about the problem by examining previous runs of the optimization algorithm and to use the obtained information to bias the search on new problem instances. The use of bias based on the results of other learning tasks is also commonplace in machine learning where it is referred to as {\em inductive transfer} or {\em transfer learning}~\cite{Pratt:91,Caruana:97}. 
 Since learning model structure is often the most computationally expensive task in model building, learning from experience often focuses on identifying regularities in model structure and using these regularities to bias structural learning in future runs. 

Analyzing probabilistic models built by hBOA and other EDAs is straightforward. The more challenging facet of implementing learning from experience in practice is that one must make sure that the collected statistics are meaningful with respect to the problem being solved. 
The key to make the learning from experience work is to ensure that the pairs of variables are classified into a set of {\em categories} so that the pairs in each category have a lot in common and can be expected to be either correlated or independent simultaneously~\cite{Pelikan:12}. This section describes one approach to doing that~\cite{Pelikan:12}, in which pairs of variables are classified into categories based on a predefined distance metric on variables.

\vspace*{-1.5ex}
\mysubsection{Distance Metric for Additively Decomposable Functions}
\vspace*{-0.5ex}
For many optimization problems, the objective function (fitness function) can be expressed as an additively decomposable function (ADF):

\vspace*{-1.8ex} 
\begin{equation}
f(X_1,\ldots,X_n) = \sum_{i=1}^m f_i(S_i),
\end{equation}
\vspace*{-1.8ex}

\noindent
where $(X_1,\ldots,X_n)$ are problem's decision variables (string positions), $f_i$ is the $i$th subfunction, and $S_i\subset \{X_1,X_2,\ldots,X_n\}$ is the subset of variables contributing to $f_i$. While there may often exist multiple ways of decomposing the problem using additive decomposition, one would typically prefer decompositions that minimize the sizes of subsets $\{S_i\}$. Note that the difficulty of ADFs is not fully determined by the order of subproblems, but also by the definition of the subproblems and their interaction; even with subproblems of order only 2 or 3, the problem can be NP-complete. 

The definition of a distance between two variables of an ADF used in this paper as well as ref.~\cite{Pelikan:12} follows the work of Hauschild et al.~\cite{Hauschild:12,Hauschild:09c,Hauschild:08}. Given an ADF, we define the distance between two variables using a graph $G$ of $n$ nodes, one node per variable. For any two variables $X_i$ and $X_j$ in the same subset $S_k$, we create an edge in $G$ between the nodes $X_i$ and $X_j$. Denoting by $l_{i,j}$ the number of edges along the shortest path between $X_i$ and $X_j$ in $G$ (in terms of the number of edges), we define the distance between two variables as

\vspace*{-1.6ex}
\[
D(X_i,X_j) = 
\left\{
\begin{array}{ll}
l_{i,j} & \mbox{if a path between $X_i$ and $X_j$ exists,}\\
n & \mbox{otherwise.}
\end{array}
\right.
\]
\vspace*{-1.8ex}

\noindent
The above distance measure makes variables in the same subproblem close to each other, whereas for the remaining variables, the distances correspond to the length of the chain of subproblems that relate the two variables. The distance is maximal for variables that are completely independent (the value of a variable does not influence the contribution of the other variable in any way). 

Since interactions between problem variables are encoded mainly in the subproblems of the additive problem decomposition, the above distance metric should typically correspond closely to the likelihood of dependencies between problem variables in probabilistic models discovered by EDAs. Specifically, the variables located closer with respect to the metric should more likely interact with each other. This observation has been confirmed with numerous experimental studies across a number of important problem domains from spin glasses distributed on a finite-dimensional lattice~\cite{Hauschild:09c,Pelikan:12} to NK landscapes~\cite{Pelikan:12}. 

\vspace*{-0.7ex}
\mysubsection{Distance-Based Bias Based on Previous Runs of hBOA}
\begin{sloppy}
This section describes the approach to learning from experience developed by Pelikan and Hauschild~\cite{Pelikan:12} inspired mainly by the work of Hauschild et al.~\cite{Hauschild:12,Hauschild:08,Hauschild:09b}. 
Let us assume a set $M$ of hBOA models from prior hBOA runs on similar problems. Before applying the bias based on prior runs in hBOA, the models in $M$ are first processed to generate data that will serve as the basis for introducing the bias. The processing starts by analyzing the models in $M$ to determine the number $s(m,d,j)$ of splits on any variable $X_i$ such that $D(X_i,X_j)=d$ in a decision tree $T_j$ for variable $X_j$ in a model $m\in M$. Then, the values $s(m,d,j)$ are used to compute the probability $P_k(d,j)$ of a $k$th split on a variable at distance $d$ from $X_j$ in a dependency tree $T_j$ given that $k-1$ such splits were already performed in $T_j$:

\vspace*{-1.3ex}
\begin{equation}
P_k(d,j) = \frac{\left|\{m\in M: s(m,d,j)\geq k\}\right|}{\left|\{m\in M: s(m,d,j)\geq k-1\}\right|}\cdot
\end{equation}
\vspace*{-1.3ex}
\end{sloppy}

\noindent
Recall that the BDe metric for evaluating the quality of probabilistic models in hBOA contains two parts: (1) the prior probability $p(B|\xi)$ of the network structure $B$, and (2) the posterior probability $p(D|B,\xi)$ of the data (population of selected solutions) given $B$. Pelikan and Hauschild~\cite{Pelikan:12} proposed to use the prior probability distribution $p(B|\xi)$ to introduce a bias based on distance-based statistics from previous hBOA runs represented by $P_k(d,j)$ by setting
\vspace*{-1.3ex}
\begin{equation}
\label{eq-pb-bias}
p(B|\xi) = c \prod_{d=1}^n \prod_{j=1}^n \prod_{k=1}^{n_s(d,j)} P^{\kappa}_k(d,j),
\end{equation}
\vspace*{-1.8ex}

\noindent 
where $n_s(d,j)$ denotes the number of splits on any variable $X_i$ in $T_j$ such that $D(X_i,X_j)=d$, $\kappa>0$ is used to tune the strength of bias (the strength of bias increases with $\kappa$), and $c$ is a normalization constant.
Since log-likelihood is typically used to evaluate model quality, when evaluating the contribution of any particular split, the change of the prior probability of the network structure can still be done in constant time. 


\mysection{Experiments}
\label{section-experiments}

\mysubsection{Test Problems and Experimental Setup}
The experiments were done for three problem classes known to be difficult for most genetic and evolutionary algorithms:
\begin{inparaenum}[(1)]
\item Three-dimensional Ising spin glasses were considered with $\pm J$ couplings and periodic boundary conditions~\cite{Pelikan:06,young1998}; two problem sizes were used, $n=6\times6\times6=216$ spins and $n=7\times7\times7=343$ spins with 1,000 unique problem instances for each $n$. 
\item Minimum vertex cover was considered for random graphs of fixed ratio $c$ of the number of edges and number of nodes~\cite{Pelikan:07b,Weigt:01}; two ratios ($c=2$ and $c=4$) and two problem sizes ($n=150$ and $n=200$) were used with 1,000 unique problem instances for each combination of $c$ and $n$. 
\item MAXSAT was considered for mapped instances of graph coloring with graphs created by combining regular ring lattices (with probability $1-p$) and random graphs (with probability $p$)~\cite{Pelikan:03*,Gent:99}; 100 unique problem instances of $n=500$ bits (propositions) were used for each considered value of $p$, from $p=2^{-8}$ (graphs nearly identical to a regular ring lattice) to $p=2^{-1}$ (graphs with half of the edges random).
\end{inparaenum}
For more information about the test problems, we refer the reader to refs.~\cite{Pelikan:06,Pelikan:07b,Pelikan:03*}.

The maximum number of iterations for each problem instance was set to the number of bits in the problem; according to preliminary experiments, this upper bound was sufficient.
Each run was terminated either when the global optimum was found, when the population consisted of copies of a single candidate solution, or when the maximum number of iterations was reached. For each problem instance, we used bisection~\cite{Pelikan:book,Sastry:01c} to ensure that the population size was within $5\%$ of the minimum population size to find the optimum in 10 out of 10 independent runs. 
Bit-flip hill climbing (HC)~\cite{Pelikan:book} was incorporated into hBOA to improve its performance on all test problems except for the minimum vertex cover; HC was used to improve every solution in the population. For minimum vertex cover, a repair operator based on ref.~\cite{Pelikan:07b} was incorporated instead.
The strength of the distance-based bias was tweaked using $\kappa\in\{1,3,5,7,9\}$.

To ensure that the same problem instances were not used for defining the bias as well as for testing it, 10-fold crossvalidation was used when evaluating the effects of distance-based bias derived from problem instances of the same size. 
For each set of problems (by a set of problems we mean a set of random problem instances generated with one specific set of parameters), problem instances were randomly split into 10 equally sized subsets. In each round of crossvalidation, 1 subset of instances was left out and hBOA was run on the remaining 9 subsets of instances. The runs on the 9 subsets produced models that were analyzed in order to obtain the probabilities $P_k(d,j)$ for all $d$, $j$, and $k$. The bias based on the obtained values of $P_k(d,j)$ was then used in hBOA runs on the remaining subset of instances. The same procedure was repeated for each subset; overall, 10 rounds of crossvalidation were performed for each set of instances. When evaluating the effects of distance-based bias derived from problem instances of {\em smaller} size, we did not use crossvalidation because in this case all runs had to be done on different problem instances (of different size). Most importantly, in every experiment, models used to generate statistics for hBOA bias were obtained from hBOA runs on {\em different} problem instances.
While the experiments were performed across a variety of computer architectures and configurations, the base case with no bias and the case with bias were always both run on the same computational node; the results of the two runs could therefore be compared against each other with respect to the actual CPU (execution) time.


To evaluate hBOA performance, we focus on the multiplicative speedup with respect to the execution time per run; the speedup is defined as a multiplicative factor by which the execution time improves with the distance-based bias compared to the base case. For example, an execution-time speedup of $2$ indicates that the bias allowed hBOA to find the optimum using only half the execution time compared to the base case without the bias. We also report the percentage of runs for which the execution time was strictly improved (shown in parentheses after the corresponding average multiplicative speedup). 

In addition to the speedups achieved for various values of $\kappa$, we examine the ability of the distance-based bias based on prior runs to apply across a range of problem sizes; this is done by using previous runs on instances of one size to bias runs on instances of another size. Since for MAXSAT, we only used instances of one size, this facet was only examined for the other two problem classes. 

Finally, we examine the combination of the distance-based bias based on prior runs and the sporadic model building~\cite{DBLP:journals/gpem/PelikanSG08}. Specifically, we apply sporadic model building on its own using the model-building delay of $\sqrt{n}/2$ as suggested by ref.~\cite{DBLP:journals/gpem/PelikanSG08}, and then we carry out a similar experiment using both the distance-based bias as well as the sporadic model building, recording the speedups with respect to the base case. Ideally, we would expect the speedups from the two sources to multiply. Due to the time requirements of solving MAXSAT, the combined effects were studied only for the remaining two problem classes.

\vspace*{-0.7ex}
\mysubsection{Results}
\vspace*{-0.48ex}

The results presented in tables~\ref{table-sg-results}, \ref{table-mvc-results} and~\ref{table-maxsat-results} confirm the observation from ref.~\cite{Pelikan:12} that the stronger the bias the greater the benefits, at least for the examined range of $\kappa\in\{1,3,5,7,9\}$ and most problem settings; that is why in the remainder of this discussion we focus on $\kappa=9$. In all cases, the distance-based bias yielded substantial speedups of about~$1.2$ to~$3.1$. Best speedups were obtained for the minimum vertex cover. In all cases, performance on at least about $70\%$ problem instances was strictly improved in terms of execution time; in most cases, the improvements were observed in a much greater majority of instances. The speedups were substantial even when the bias was based on prior runs on problem instances of different, smaller size; in fact, the speedups obtained with such a bias were nearly identical to the speedups with the bias based on the instances of the same size. The results thus provide clear empirical evidence that the distance-based bias is applicable even when the problem instances vary in size, which was argued~\cite{Pelikan:12} to be one of the main advantages of the distance-based bias over prior work in the area but was not demonstrated. Finally, the results show the nearly multiplicative effect of the distance-based bias and sporadic model building, providing further support for the importance of the distance-based bias; the combined speedups ranged from about 4 to more than 11.

\input{tables}

\mysection{Summary and Conclusions}
\label{section-conclusions}
This paper extended the prior work on efficiency enhancement of the hierarchical Bayesian optimization algorithm (hBOA) using a distance-based bias derived from prior hBOA runs~\cite{Pelikan:12}.  
The paper demonstrated that (1) the distance-based bias yields substantial speedups on several previously untested classes of challenging, NP-complete problems, (2) the approach is applicable even when prior runs were executed on problem instances of different size, and (3) the approach can yield nearly multiplicative speedups when combined with other efficiency enhancement techniques. In summary, the results presented in this paper together with the prior work~\cite{Pelikan:12} provide clear evidence that learning from experience using a distance-based bias has a great potential to improve efficiency of hBOA in particular and estimation of distribution algorithms (EDAs) in general.

Several topics are of central importance for future work. 
The approach should be adapted to other model-directed optimization techniques, including other EDAs
and genetic algorithms with linkage learning. The approach should also be modified to introduce
bias on problems that cannot be formulated using an additive decomposition in a straightforward
manner or such a decomposition is not practical. 
Finally, it is important to study the limitations of the proposed approach, and create
theoretical models to automatically tune the strength of the bias and predict expected speedups.

\section*{Acknowledgments} 
This project was sponsored by the National Science Foundation under grants ECS-0547013 and IIS-1115352, and by the Univ. of Missouri--St. Louis through the High Performance Computing Collaboratory sponsored by Information Technology Services. Most experiments were performed on the Beowulf cluster maintained by ITS at the Univ. of Missouri in St. Louis and the HPC resources at the University of Missouri Bioinformatics Consortium. Any opinions, findings, and conclusions or recommendations expressed in this material are those of the authors and do not necessarily reflect the views of the National Science Foundation.

\vspace*{-1.5ex}
\bibliographystyle{splncs}
\begin{small}
\bibliography{mybib}
\end{small}

\end{document}

%% file: tables.tex

\begin{table}[t]
\floatbox{table}[\textwidth]{
\caption{Results for 3D spin glass.}
\label{table-sg-results}
\begin{subfloatrow}
\subfloat[Results for 10-fold crossvalidation with priors from other
instances of the same size.]
{\begin{tabular}{|c|cc|}\hline
\multirow{2}{*}{$\kappa$} & \multicolumn{2}{|c|}{CPU speedup}\\
& $n=216$ & $n=343$ \\\hline
1 & $0.40$ $(~0\%)$ & $0.43$  $(~0\%)$ \\
3 & $1.00$ $(43\%)$  & $1.08$  $(60\%)$ \\
5 & $1.23$  $(71\%)$ & $1.32$  $(85\%)$ \\
7 & $1.24$  $(70\%)$ & $1.34$  $(81\%)$ \\
9 & $1.21$  $(66\%)$ & $1.20$  $(67\%)$ \\
\hline
\end{tabular}
}
\subfloat[Results for $n=343$ with priors based on models obtained on
problem instances of {\em smaller}  size, $n=216$.]
{~~~\begin{tabular}{|c|c|}\hline
\multirow{2}{*}{$\kappa$} & \multirow{2}{*}{CPU speedup}\\
&\\\hline
1 & $0.43$ $(~1\%)$ \\ 
3 & $1.05$ $(61\%)$ \\ 
5 & $1.33$ $(85\%)$ \\ 
7 & $1.34$ $(82\%)$ \\ 
9 & $1.26$ $(75\%)$ \\ 
\hline
\end{tabular}
~~~
}
~~
\subfloat[Results for a combination of distance-based bias (DBB)
and sporadic model building (SMB) for $n=343$. 10-fold
crossovalidation was used.]
{\begin{tabular}{|c|cc|}\hline
\multirow{2}{*}{$\kappa$} & \multicolumn{2}{|c|}{CPU speedup}\\
& ~DBB+SMB & ~SMB~~\\\hline
1 & 
$1.85$ $(99\%)$ & $3.20$ $(99\%)$ \\ 
3 & 
$3.29$ $(99\%)$ & $3.20$  $(99\%)$ \\ 
5 & 
$4.04$ $(99\%)$ & $3.20$  $(99\%)$ \\ 
7 & 
$4.23$ $(99\%)$ & $3.20$  $(99\%)$ \\ 
9 & 
$4.03$ $(99\%)$ & $3.20$  $(99\%)$ \\ 
\hline
\end{tabular}
}
\end{subfloatrow}
}


\end{table}



\begin{table}[h]
\floatbox{table}[\textwidth]{
\caption{Results for minimum vertex cover.}
\label{table-mvc-results}
\begin{subfloatrow}
\subfloat[Results for 10-fold crossvalidation with priors from other instances of the same size.]
{
\begin{tabular}{|c|cc|}\hline
\multicolumn{3}{|c|}{$c=2$}\\\hline
\multirow{2}{*}{$\kappa$} & \multicolumn{2}{|c|}{CPU speedup}\\
& $n=150$ & $n=200$ \\\hline
1 & $0.57$ $(~2\%)$ & $0.45$ $(~0\%)$\\
3 & $1.95$  $(91\%)$  & $1.63$ $(87\%)$\\
5 & $2.78$ $(96\%)$ & $2.69$ $(94\%)$ \\
7 & $3.04$ $(95\%)$ & $2.98$ $(94\%)$\\
9 & $3.10$ $(93\%)$ & $2.95$ $(92\%)$ \\
\hline
\end{tabular}
}
\subfloat[Results for $n=200$ with priors based on models obtained on
problem instances of {\em smaller} size, $n=150$.]
{~~~
\begin{tabular}{|c|c|}\hline
\multicolumn{2}{|c|}{$c=2$}\\\hline

\multirow{2}{*}{$\kappa$} & \multirow{2}{*}{CPU speedup}\\
&\\
\hline
1 & $0.53$ $(~2\%)$ \\
3 & $1.95$ $(91\%)$ \\
5 & $2.79$ $(95\%)$ \\
7 & $2.99$ $(94\%)$ \\
9 & $3.02$ $(91\%)$ \\ 
\hline
\end{tabular}
~~~}~~
\subfloat[Results for a combination of distance-based bias (DBB) and sporadic model building (SMB) for
$n=200$. 10-fold crossovalidation was used.]
{
\begin{tabular}{|c|cc|}\hline
\multicolumn{3}{|c|}{$c=2$}\\\hline
\multirow{2}{*}{$\kappa$}& \multicolumn{2}{|c|}{CPU speedup}\\
& ~DBB+SMB~ & ~SMB~ \\\hline
1 & 
~$3.12$ $(~99\%)$ & $4.89$ \\ 
3 & 
~$6.89$ $(100\%)$& $4.89$ \\ 
5 & 
$10.25$ $(100\%)$& $4.89$ \\ 
7 & 
$11.38$ $(100\%)$& $4.89$ \\ 
9 & 
$11.29$ $(~99\%)$& $4.89$ \\ 
\hline
\end{tabular}
}
\end{subfloatrow}
}


~~
\begin{tabular}{|c|cc|}\hline
\multicolumn{3}{|c|}{$c=4$}\\\hline
\multirow{2}{*}{$\kappa$} & \multicolumn{2}{|c|}{CPU speedup}\\
& $n=150$ & $n=200$ \\\hline
1 & $0.28$ $(~0\%)$ & $0.17$ $(~0\%)$ \\
3 & $0.97$  $(39\%)$ & $0.53$ $(~4\%)$ \\
5 & $1.56$  $(82\%)$ & $1.16$ $(62\%)$ \\
7 & $1.97$  $(88\%)$ & $1.65$ $(81\%)$ \\
9 & $2.27$  $(89\%)$ & $1.91$ $(85\%)$ \\
\hline
\end{tabular}~~~~~
~~\begin{tabular}{|c|c|}\hline
\multicolumn{2}{|c|}{$c=4$}\\\hline
\multirow{2}{*}{$\kappa$} & \multirow{2}{*}{CPU speedup}\\
&\\
\hline
1 & $0.23$ $(~0\%)$ \\ 
3 & $0.86$  $(27\%)$\\ 
5 & $1.50$  $(79\%)$\\ 
7 & $1.89$  $(85\%)$\\ 
9 & $2.12$  $(84\%)$\\ 
\hline
\end{tabular}~~~~~~~
\begin{tabular}{|c|cc|}\hline
\multicolumn{3}{|c|}{$c=4$}\\\hline
\multirow{2}{*}{$\kappa$} & \multicolumn{2}{|c|}{CPU speedup}\\
& ~DBB+SMB~ & ~SMB~ \\\hline
1 & 
$1.88$ $(~82\%)$ & $4.54$ \\ 
3 & 
$3.24$ $(~96\%)$& $4.54$ \\ 
5 & 
$5.00$ $(~99\%)$ & $4.54$ \\ 
7 & 
$6.15$ $(~99\%)$ & $4.54$ \\ 
9 & 
$6.60$ $(~99\%)$ & $4.54$ \\ 
\hline
\end{tabular}~~

\end{table}


\begin{table}[h]
\caption{Results for MAXSAT.}
\label{table-maxsat-results}

\centering

\begin{tabular}{|c|cccc|}\hline
\multirow{2}{*}{$\kappa$} & \multicolumn{4}{|c|}{CPU speedup}\\
& ~$p=2^{-1}$ & ~$p=2^{-2}$~ & ~$p=2^{-4}$~ & ~$p=2^{-8}$~ \\\hline
1 & $0.13$ $(~0\%)$ & $0.22$ $(~0\%)$ & $0.22$ $(~0\%)$ & $0.38$ $(~~0\%)$ \\
3 & $0.41$ $(~0\%)$ & $0.53$ $(~0\%)$ & $0.48$ $(~0\%)$ & $1.01$ $(~49\%)$ \\
5 & $0.81$ $(25\%)$ & $0.82$ $(18\%)$ & $0.74$ $(~4\%)$ & $1.63$ $(100\%)$ \\
7 & $1.38$ $(69\%)$ & $1.09$ $(55\%)$ & $1.03$ $(54\%)$ & $1.84$  $(100\%)$ \\
9 & $2.31$ $(94\%)$ & $1.38$ $(81\%)$ & $1.28$ $(89\%)$ & $1.90$  $(100\%)$ \\
\hline
\end{tabular}

\end{table}
